# BRIEF: BRain-Inspired network connection search with Extensive temporal feature Fusion enhances disease classification

Xiangxiang Cui[1], Min Zhao[2], Dongmei Zhi[1], Shile Qi[3], Vince D Calhoun[4], Jing Sui[1*]

[1]McGovern Brain Imaging Institute, State Key Laboratory of Cognitive Neuroscience and Learning, Beijing Normal University, Beijing, China

[2] Department of Computer Science and Technology, Tsinghua University, Beijing, China

[3]College of Artificial Intelligence, Nanjing University of Aeronautics and Astronautics, Nanjing, China, and The Key Laboratory of Brain-Machine Intelligence Technology, Ministry of Education, Nanjing University of Aeronautics and Astronautics, Nanjing, China

[4]Tri-institutional Center for Translational Research in Neuroimaging and Data Science (TReNDS), Georgia Institute of Technology, Emory University and Georgia State University, Atlanta, Georgia, United States

**\*Corresponding Author: Jing Sui PhD, Professor**

Email: jsui@bnu.edu.cn

State Key Laboratory of Cognitive Neuroscience and Learning, Beijing Normal University, China

# Abstract

Existing deep learning models for functional MRI-based classification have limitations in network architecture determination (relying on experience) and feature space fusion (mostly simple concatenation, lacking mutual learning). Inspired by the human brain's mechanism of updating neural connections through learning and decision-making, we proposed a novel BRain-Inspired feature Fusion (BRIEF) framework, which is able to optimize network architecture automatically by incorporating an improved neural network connection search (NCS) strategy and a Transformer-based multi-feature fusion module. Specifically, we first extracted 4 types of fMRI temporal representations, i.e., time series (TCs), static/dynamic functional connection (FNC/dFNC), and multi-scale dispersion entropy (MsDE), to construct four encoders. Within each encoder, we employed a modified Q-learning to dynamically optimize the NCS to extract high-level feature vectors, where the NCS is formulated as a Markov Decision Process. Then, all feature vectors were fused via a Transformer, leveraging both stable/time-varying connections and multi-scale dependencies across different brain regions to achieve the final classification. Additionally, an attention module was embedded to improve interpretability. The classification performance of our proposed BRIEF was compared with 21 state-of-the-art models by discriminating two mental disorders from healthy controls: schizophrenia (SZ, n=1100) and autism spectrum disorder (ASD, n=1550). BRIEF demonstrated significant improvements of 2.2% to 12.1% compared to 21 algorithms, reaching an AUC of 91.5% ± 0.6% for SZ and 78.4% ± 0.5% for ASD, respectively. To the best of our knowledge, this is the first attempt to incorporate a brain-inspired, reinforcement learning strategy to optimize fMRI-based mental disorder classification, showing significant potential for identifying precise neuroimaging biomarkers.

# Introduction

Deep learning for neuroimaging has seen a proliferation of various feature extraction approaches designed to study functional magnetic resonance imaging (fMRI) activity, aiming to gain a deeper understanding of neural mechanisms underlying psychiatric disorders [1]. For example, various approaches have been developed to utilize either functional network connectivity (FNC), time courses (TC), or dynamic functional network connectivity (dFNC), individually or in combination, thereby aiding in the diagnosis of mental disorders [2-4]. To fully leverage the abundant information contained in FNC, [5] proposed a deep neural network (DNN) with layer-wise relevance propagation (LRP), to distinguish SZ from healthy controls (HCs) using FNC. By conducting LRP, they identified the FNC patterns that exhibit the highest discriminative power in SZ classification. Zhang et al. proposed A-GCL [6], which employs graph contrastive learning with edge-dropped graphs to extract features in fMRI-derived brain networks. Furthermore, the TCs capture dynamic neural activities. Yan et al. [3] developed the Multi-scale RNN (MsRNN), which combines CNN and RNN to enhance fMRI classification by integrating spatio-temporal features. [2] proposed a hybrid deep learning framework aimed at boosting classification accuracy. Recent work has further advanced this field with iTransformer [7] which inverts the roles of attention mechanisms and feed-forward networks to capture multivariate correlations in time series. PatchTST [8] segments time series into subseries-level patches as transformer input tokens, improving long-term forecasting accuracy. What's more, ACIFBN [9] (Asynchronous Common and Individual Functional Brain Network) is used to model asynchronous functional interactions for improved Alzheimer's disease diagnosis. Integrating spatiotemporal information from both FNC and TC studies has shown a substantial enhancement in the accuracy of psychiatric disorder classification. However, the specific neural architecture optimization typically relies on manual trial and error rather than an automated approach. Prior work also often performs feature fusion [2] via logistic regression which is likely suboptimal.

We can consider neural network optimization in light of the behaviorist theory in psychology, especially Skinner's operant conditioning (operant conditioning) [10], which suggests that the behavior of animals (including humans) is shaped by the effects of rewards and punishments. Inspired by this, we proposed a brain-inspired network connection search (NCS) using Q-learning. By mapping different modules of Q-learning to the optimization of network connections, we established the NCS framework for automatic network connection optimization. NCS has many similarities to the learning mechanisms in the human brain. This process can be considered analogous to the decision-making mechanisms [11] in the human brain, especially in regions involved in reward prediction and value evaluation, such as the prefrontal cortex [11, 12] and the basal ganglia [13]. Specifically: 1) NCS uses Q-values to predict the cumulative reward of new network connections, which is similar to the human brain's ability to predict the outcomes of future behaviors [14]. 2) Both learn through trial and error. Humans try different behaviors and observe the results [15], while NCS continuously tries new connections and updates the Q-values to determine better outcomes. 3) Meanwhile, NCS's Q-values are gradually adjusted by considering both the current connection reward and the future connection reward estimate, similar to how the human brain's dopamine system adjusts dopamine release or reduction based on prediction errors, thereby affecting future decisions [16]. 4) The discount factor (gamma) in the Q-values represents the value of future connection rewards relative to immediate connection rewards. This reflects the natural human tendency to balance between immediate gratification and delayed gratification [17]. 5) When faced with new situations, humans balance the possibility of exploring the unknown with exploiting the best-known option [18]. NCS also achieves this balance through methods such as the epsilon-greedy strategy. Humans can sacrifice short-term interests for long-term goals, and NCS can simulate similar long-term planning abilities by recursively calculating cumulative connection rewards. 6) The brain operates through complex multi-scale connection networks, integrating information from various spatial and temporal scales [19-21]. Some brain science studies emphasize the importance of brain connectivity in understanding neuropsychiatric disorders [19] [20, 21]. These studies collectively stress the importance of understanding brain connectivity patterns. Therefore, long and short-range connections have the potential to play a greater role in the diagnosis of psychiatric disorders. In addition, existing studies [22, 23] highlighted the importance of multiscale approaches in brain disorder research. For instance, [23] developed a multi-Scale multi-hop GCN to estimate fluid intelligence score by using FNC. From the results of NCS optimization, long and short-range connections, as well as residual connections and concatenate connections, together form a complex multi-scale connection network. In addition, unlike traditional neural architecture search (NAS) [24] with its large search space and high computational costs, NCS optimizes within advanced network architectures, streamlining the search and enhancing efficiency and precision at reduced computational costs.

For features fusion, we extract multiple fMRI features and then these features output feature vectors after passing through the feature extraction network. Additional feature fusion occurs after concatenating the feature vectors: 1) Firstly, TCs not only can compute FNC, but also yield dynamic FNC (dFNC) and multi-scale dispersion entropy (MsDE). Collectively, these features reflect the multi-scale information embedded in fMRI data, which is instrumental in furthering our understanding of the role of fMRI in the diagnosis of mental disorders. Using various-sized sliding windows on fMRI time series data, we extract FNC, dFNC, and MsDE from TCs, enhancing our understanding of both static and dynamic brain connectivity. FNC uses a large window for static views, while dFNC uses smaller windows to capture dynamics, significantly improving psychiatric disorder classification [4, 25, 26]. We integrated multi-scale dispersion entropy (MsDE) into our analysis of fMRI temporal data, drawing inspiration from the methodologies described in reference [22]. MsDE explores brain signal complexity across multiple scales, effectively capturing dynamic brain activity. 2) Secondly, these features are then input into the extensive temporal features extraction module in **Fig. 1**, and then four feature vectors will be output. 3) Finally, while [2] constructs a two features extraction network, relying solely on the logistic regression (LR) model for feature fusion may constrain the ability to fully and efficiently integrate these diverse features. We introduce a transformer [27] to enhance feature fusion in fMRI analysis, aiming for more effective classification of psychiatric disorders (see **Fig. 1**). Therefore, four feature vectors are spliced and input into a transformer for further integration to improve classification performance.

In summary, building on the points above, we developed a framework BRIEF based on a brain-inspired network connection search with extensive temporal feature fusion, which significantly improved the classification of mental disorders. To validate the effectiveness of our proposed BRIEF, we employed multi-site fMRI datasets from studies on schizophrenia and autism. The results achieved an accuracy improvement ranging from 2.2% to 12.1% compared to a multitude of existing methods. Specifically, NCS, as a key component of proposed BRIEF, optimizes the feature extraction network, enabling more precise and efficient extraction of extensive temporal fMRI features. By integrating the attention mechanism, our framework identified the most discriminative brain regions for schizophrenia (SZ), primarily located in the striatum and precentral gyrus. This discovery paves the way for providing potential neuroimaging biomarkers to aid in the diagnosis of psychiatric disorders.

## Materials and Methods

All participants provided informed consent approved by the Institutional Review Board (IRB) at each recruiting site before any data were collected. All data and clinical assessment used for current study were anonymized. The use of human subjects in our proposed research was approved by Beijing Normal University as an exempt study.

### A. Participants and Feature Extraction

This analysis utilized two datasets: a proprietary schizophrenia database (558 patients and 542 healthy controls) and the publicly available Autism Brain Imaging Data Exchange (ABIDE) (743 individuals with autism and 779 healthy controls). Demographic characteristics of all participants are summarized in **Table 1**, with between-group comparisons of age and gender conducted using two-sample t-tests and chi-square tests, respectively. The proprietary data were collected from seven clinical institutions across different regions using various imaging equipment. Functional MRI preprocessing was performed using SPM8 software (http://www.fil.ion.ucl.ac.uk/spm/) with a pipeline consisting of slice timing correction, motion correction, normalization to Montreal Neurological Institute (MNI) standard space (3×3×3 mm³ voxel resampling), and noise reduction with spatial smoothing using an 8 mm FWHM Gaussian kernel.

Table 1 Demographic information for the In-House SZ and ABIDE datasets

| Dataset | Group | Subject | Gender | Age |
|---|---|---|---|---|
| **In-House SZ** | SZ | 558 | 292/266 | 27.6±7.1 |
|  | HC | 542 | 276/266 | 28.0±7.2 |
|  | p | NA | 1.9e-06 | 0.06 |
| ABIDE | Autism | 743 | 645/98 | 16.6±9.1 |
|  | HC | 779 | 585/194 | 16.2±8.4 |
|  | p | NA | 5.6e-09 | 0.28 |

Notes: p-value: the significance value of Chi-square test for gender and two sample t-test for age. NA: not applicable.

We employed independent component analysis (ICA) using the GIFT software (http://trendscenter.org/software/gift) platform to process functional neuroimaging data and extract spatial functional networks (Fig. 1a). To achieve fine-grained network segregation, we adopted a high model order approach with 100 components. At the subject level, 150 features were retained through principal component analysis (PCA); subsequently, at the group level, PCA dimensionality reduction to 100 components was performed. To ensure result stability, we utilized the ICASSO framework to repeat the Infomax ICA algorithm 20 times, selecting the most representative solution, which yielded 100 stable group independent components. After further screening and evaluation, we ultimately identified and characterized 50 intrinsic connectivity networks with biological significance.

Subsequently, we reconstructed individual-specific temporal dynamics and spatial patterns using the GIG-ICA approach. The temporal components underwent comprehensive post-processing, including multi-order (linear, quadratic, and cubic) detrending, nuisance regression of head motion parameters and their first derivatives, removal of signal spikes, and temporal filtering with a low-pass filter. Group-level temporal dynamics of the independent components are illustrated in the supplementary materials (**Fig. S4**). Functional connectivity between independent components was quantified through Pearson correlation coefficients, which were normalized using Fisher's Z-transformation. This analytical pipeline yielded a symmetric 50×50 functional network connectivity matrix (FNC) for each individual participant. Group-level connectivity patterns and between-group differences are presented in the supplementary materials (**Fig. S2**).

Dynamic functional network connectivity (dFNC) was computed using a sliding window approach with window size is 10 timepoints and step size is 1 applied to the time courses (TCs). Within each window, Pearson correlation coefficients between all component pairs were calculated, and the upper triangular elements were extracted to form feature vectors. The temporal variability patterns of dFNC across groups are illustrated in the supplementary materials (**Fig. S1**).

Multi-scale dispersion entropy (MsDE) was computed to characterize temporal complexity patterns across multiple scales applied to the time courses (TCs). For each scale factor $\tau$ ranging from 1 to 80, signals were coarse-grained by averaging non-overlapping windows of size $\tau$. The coarse-grained signals were normalized using min-max normalization to [0, 1] range and discretized into $c=10$ classes. Embedding vectors of dimension $m=2$ with time delay $\tau=1$ was constructed from the symbolic sequences, and dispersion entropy was calculated using Shannon entropy applied to the pattern distribution. This procedure yielded 80-dimensional MsDE feature vectors for each of the 50 independent components (**Fig. S3**).

### B. Research design

As illustrated in **Fig. 1**, the proposed framework comprises three core modules: **a) Brain-Inspired Network Connection Search (NCS):** This module formulates neural architecture optimization as a Markov Decision Process, employing Q-learning to dynamically refine inter-layer connections within a manually designed backbone network (switch from original network to Graph). The NCS algorithm explores biologically plausible connection patterns, including but not limited to residual and concatenate connections, thereby enhancing the network's ability to capture spatiotemporal dependencies in fMRI data. This process mirrors neural plasticity mechanisms observed in biological systems. Compared to Neural Architecture Search (NAS), NCS is not optimized from scratch, which saves an amount of time. **b) Multiple Feature Extraction**: The framework processes preprocessed fMRI data through GIG-ICA to derive four distinct feature categories: time courses (TCs), static functional network connectivity (FNC), dynamic FNC computed with sliding windows of 160 time points, and multiscale dispersion entropy (MsDE). Each feature type undergoes dedicated encoding via NCS-optimized sub-networks, generating dimensionally feature vectors for subsequent fusion. **c) Attention-Based Feature Fusion**: A Transformer module with multi-head self-attention mechanisms integrates the extracted feature vectors. This design capitalizes on the Transformer's ability to model long-range dependencies across heterogeneous features, while avoiding its computational inefficiency when applied directly to raw fMRI data. The fused representation is ultimately fed into a classification head for final prediction.

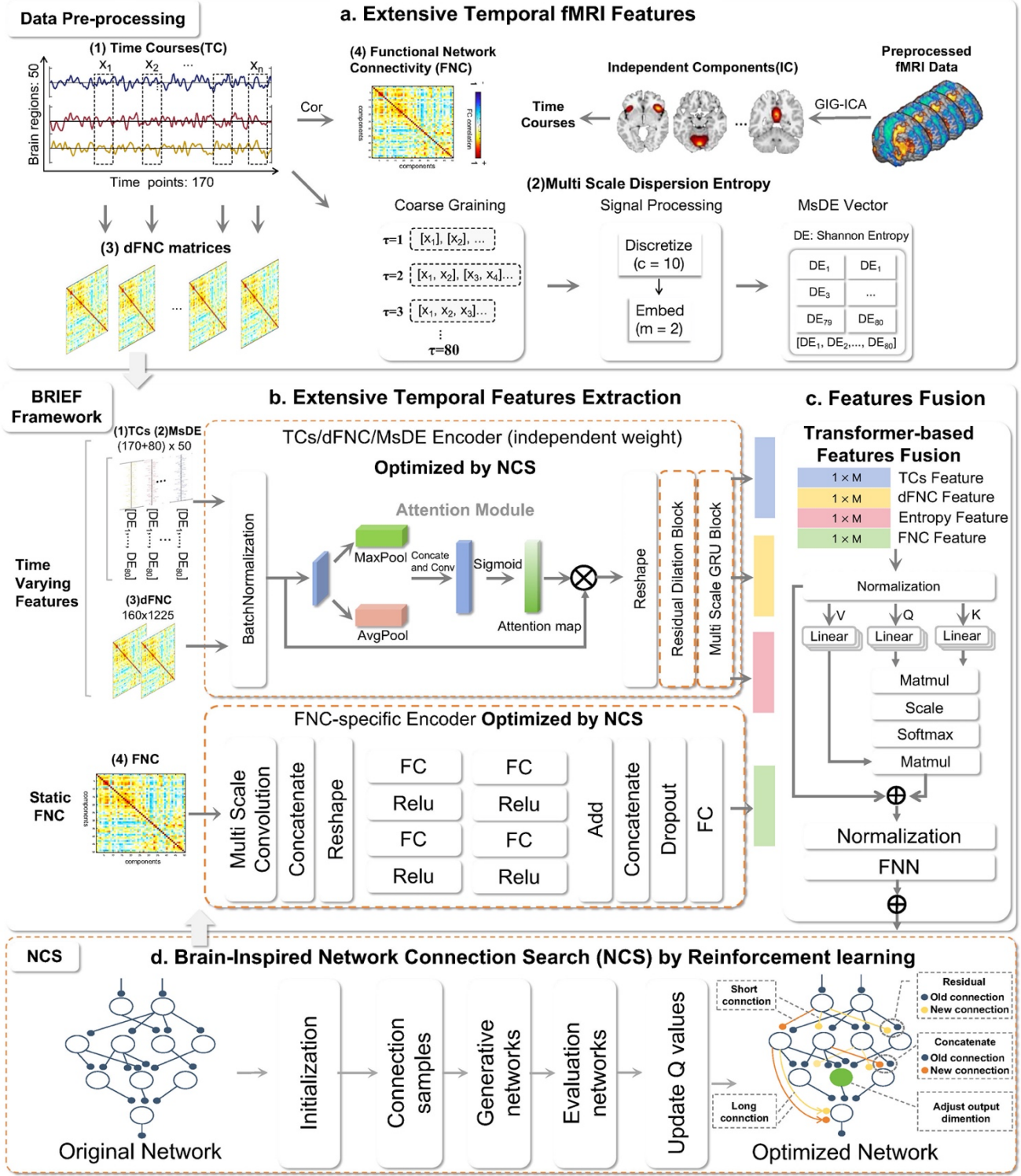

**Figure. 1** The proposed BRIEF framework for brain disorder classification. The framework consists of three main modules: 1) Data pre-processing: (a) Extensive temporal data generation: generating independent components and time courses (TCs) via GIG-ICA from preprocessed fMRI, then deriving Functional Network Connectivity (FNC), Dynamic Functional Network Connectivity (dFNC), and multi-scale dispersion entropy (MsDE) from TCs. 2) BRIEF Framework: (b) Extensive temporal feature extraction: employing specialized encoders for TCs/dFNC/MsDE (featuring attention modules, residual-dilation blocks, and multi-scale GRU blocks) and a separate encoder for FNC, all optimized by reinforcement learning-based brain-inspired network connection search (NCS) shown in (d). (c) Feature fusion: using a transformer-based approach to integrate the feature vectors from all encoders for classification.

The encoder module includes four encoders for TCs/dFNC/MsDE and FNC respectively. (a) For the TCs/dFNC/MsDE encoder, which mainly include two important blocks: the residual dilation block (**Fig. 2(A)**) and the multi-scale GRU block (**Fig. 2(B)**) are optimized by NCS. The encoder can effectively capture the temporal dynamics of brain activity. (b) For the FNC-specific encoder, which is derived from the connection optimization of exists SOTA networks using NCS. It can model the global high-order functional coherence. It's worth noting that this may be the first work to utilize reinforcement learning (**Fig. 3**) for optimizing connections on the SOTA network. This could potentially open up new avenues for future research and development in the area of network architecture and optimization. A transformer-based fusion method was used to combine multiple feature vectors from the TCs/ dFNC/ MsDE and FNC encoders output and perform classification.

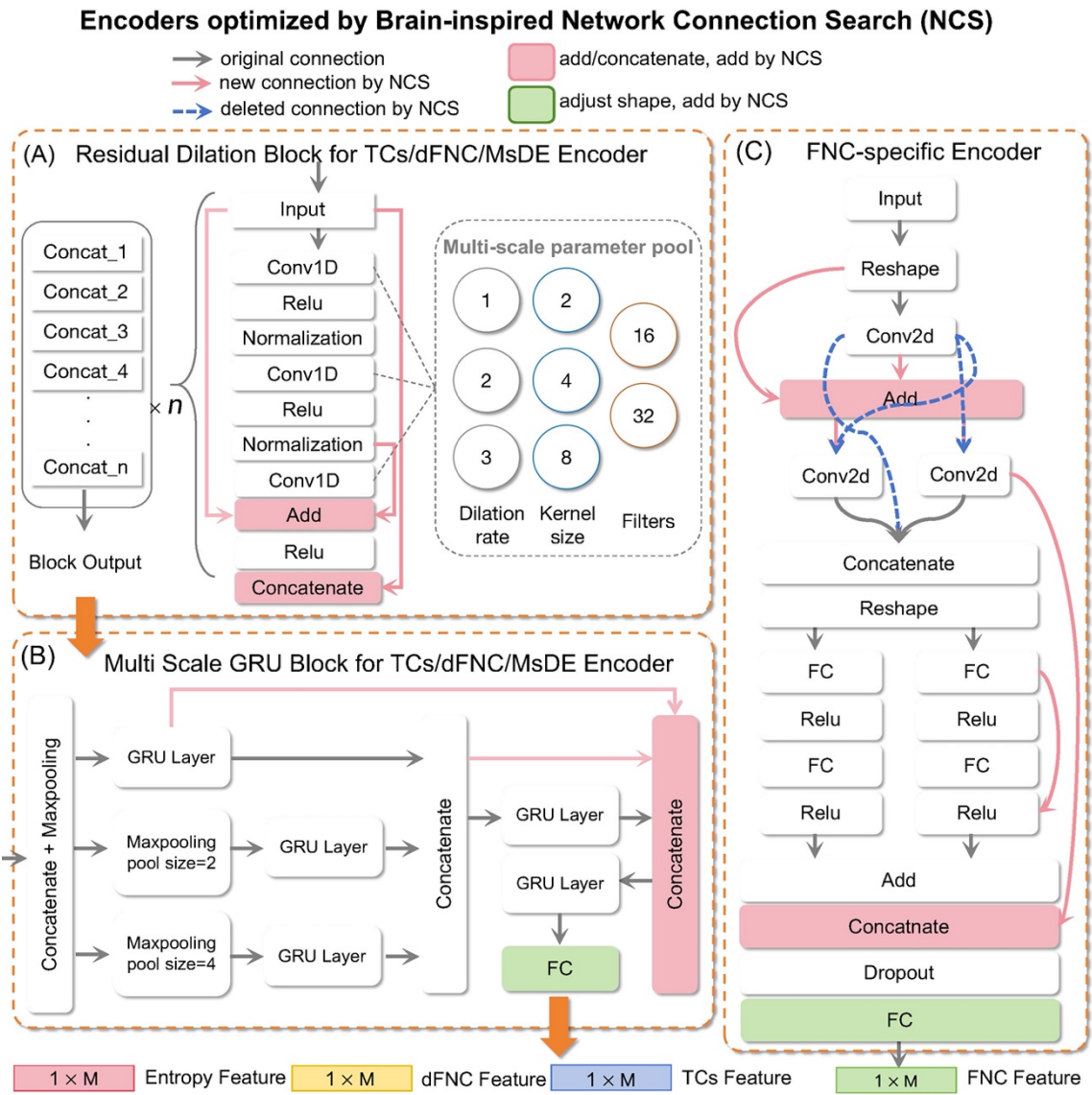

**Figure. 2** The model block optimized by brain-inspired network connection search (NCS). (A) Residual dilation block, which stacks multiple network layers according to different combinations of parameters. Three connections were added through NCS. (B) Multi-scale GRU block, which includes different scale input and multi-scale connection added by NCS. (A) and (B) blocks serve as core components of the TCs/dFNC/MsDE encoder. (C) FNC-specific encoder, whose long and short connections, add and concatenate layers as well as the last layer of shape are optimized by NCS.

Let $X = \{x_i\}_{i=1}^{H}$, $Z = \{z_i\}_{i=1}^{H}$, $P = \{p_i\}_{i=1}^{H}$ and $E = \{e_i\}_{i=1}^{H}$ denote the input FNC, TC, dFNC and multi-scale dispersion entropy features, where $x_i \in R^{B \times B}$, $z_i \in R^{\Phi \times B}$, $p_i \in R^{(\frac{\Phi-W}{S}+1) \times \frac{B(B-1)}{2}}$, $e_i \in R^{\frac{\Phi}{\sigma} \times B}$, $H$ is the number of subjects, $B$ is the number of $ICs$ (brain regions) and $\Phi$ is the number of time points, $W$ is the sliding window size, $S$ is the sliding step, $\sigma$ is defined in Section 2 (Multi-scale dispersion entropy). We denote the corresponding label set as $Y = \{y_i\}_{i=1}^{H}$, where $y_i \in \{1, \cdots C\}$ and $C = 2$.

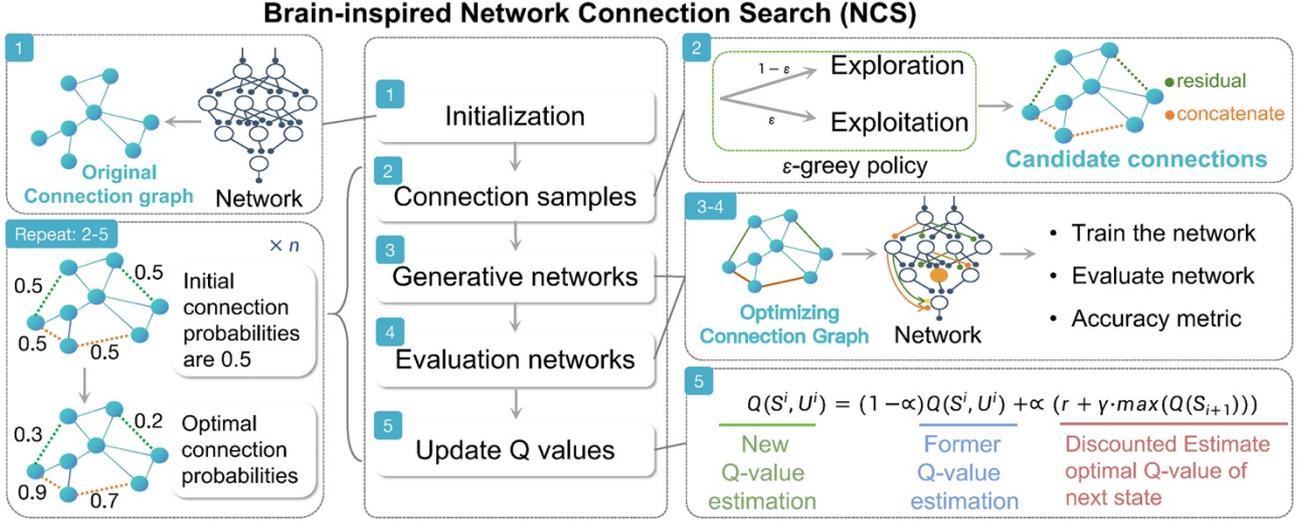

**Figure. 3** The framework of brain-inspired network connection search (NCS) which is based on Q-learning and consists of five steps: 1) Initialization: Converts the model into a graph, where each node includes layer names, types, channels, and other structural aspects that define the model architecture. 2) Connection Samples: Based on an epsilon-greedy policy, generate candidate connections and record them on the graph. 3) Generative Networks: Transform the graph into networks containing candidate connections. 4) Evaluation Networks: Train and assess the networks. 5) Update Q-values: Utilize the Bellman equation in Q-learning.

**1) Brain-inspired network connection search**

NAS [28] has become a key technology for the automated design of efficient neural networks. NAS aims to discover optimized network architectures through an automated process, greatly promoting the development and innovation of complex models. However, NAS faces several major challenges: a) High computational resource demands. b) Time-consuming search. c) Complex search space design. Our approach is inspired by the existing brain connectivity [19-21] and network optimization strategies, especially the innovations in network connections by UNet++ [29] and ResNet [30], which have significantly improved network performance through enhanced connections. We proposed brain-inspired network connection search (NCS), which focuses on optimizing potential new connections within existing SOTA networks rather than the overall structure. By identifying and refining these new connections, NCS aims to improve search efficiency, reduce computational resource requirements, and shorten search time. This method not only challenges the limitations of traditional NAS but also provides a more precise and efficient path for future network design.

Specifically, we formulate NCS as a Markov decision process (MDP). In this framework, each 'state' corresponds to the current architecture of the network's connections, while 'actions' refer to potential modifications to these connections. The 'rewards' are determined based on the performance improvements resulting from these modifications. Transitions between states occur when an action modifies the network architecture. By employing this method, we overcome the limitations of traditional NAS and offer a more precise and efficient approach to network design.

**1.1) Initialization:** First, we transform the input model into a graph representation. Next, we search the graph for all potential connections, forming a set of possible edges between nodes. Then, we initialize the Q-table in Q-learning, setting initial values for each possible state-action pair, which is a crucial preparatory step for subsequent NCS processes. The following are more detailed descriptions.

**Connection Graph:** To facilitate the analysis of potential connections within a model, a graph structure is employed. Each node represents a layer, while edges represent the flow of data between layers. The properties of each node include inputs, outputs, types of activation functions, kernel sizes, stride lengths, padding strategies, and their interconnections with other layers. Specifically, let $M$ be a deep learning model consisting of a series of layers $L = \{l_1, l_2, \ldots, l_n\}$. We define a function $f: M \to G$ that converts the model $M$ to a graph $G = (V, E)$, where $V$ is a set of nodes, with each node $V_i$ corresponding to a $i^{th}$ layer $l_i$ in $L$. $E$ is a set of edges, with each edge $E_{i,j}$ representing a connection between node $V_i$ and $V_j$. Here, $i$ and $j$ are indices in $\{1,2,\ldots,n\}$, where $n$ represents the number of layers in $L$.

**Identify all candidate connections:** Based on the graph $G$, we can analyze all potential new connections. For any two nodes $V_i$ and $V_j$, if there is no direct edge $E_{i,j}$ in $G$, we can add new edge $E_{i,j}^{new}$ from $V_i$ to $V_j$. By doing so, we ultimately generate a new graph $G'$ that includes all those potential connections.

**Initialize the Q-table:** We use Q-learning reinforcement learning to optimize new connections in the $G'$. The first step is to initialize the Q table, which is the basis for evaluating the potential utility of the new connection. Specifically, all new connections in $G'$ are actions in the Q-table. The initialization of the Q-table involves assigning an initial probability value to each action. All potential connections, representing actions in the Q-table, in graph $G'$ are initially set to a uniform probability of 0.5 at the start. This process can be expressed as:

$$Q(V_i, V_j) = 0.5, \forall i,j \in 1,2,\ldots,t, j > i, t \leq n \tag{1}$$

Where $n$ is the number of layers. $t$ is the index of feature fusion layer, which output feature vector to the Transformer. The function $Q$ denotes the action probability value in the Q-table of Q-learning. The above formula means that all new connections have an initial action probability value of 0.5. $Q(V_i,\cdot)$ represents the set of probability values for all edges originating from node $V_i$. $Q(\cdot,V_j)$ represents the set of probability values for all edges terminating at the node $V_j$. It is worth noting that the connections in the Q-table are all new connections, $(V_i, V_j)$ is $E_{i,j}^{new}$.

**1.2) Connection samples:** After above initialization, we use the $\varepsilon - greedy$ strategy to optimize action probability values of the new connections in Q-table.

Specifically, $\varepsilon$ is a parameter used to balance exploration and exploitation. Exploration involves employing unoptimized actions, while exploitation involves employing optimized actions. Initially, $\varepsilon$ is high to encourage exploration. As the algorithm progresses, $\varepsilon$ gradually decreases, which increases the proportion of optimized actions in exploitation module. Specifically, the $\varepsilon$ decreases as the number of iterations increases.

For each optimization, the algorithm will perform operations such as selecting actions, updating the status and action sequence. This process is modeled as follows: $k_{max}$ is user-defined upper limit for new connections. $S$ represents the list of starting nodes for new connections, $U$ represents the list of ending nodes for these connections, and $\Upsilon$ represents the type (concatenation or residual) for these connections. $S$ is initialized with a specific custom node, $U$ and $\Upsilon$ are initialized as empty lists. The following $k \in \{1,2,\ldots,k_{max}\}$. To be specific:

**Connection selection**

$$Nod_k = \begin{cases} Argmax(Q(S[-1]),\cdot), & if\ a_k > \varepsilon, \\ RandomChoice(Q(S[-1],\cdot)), & otherwise. \end{cases} \quad (2)$$

At each $k$, the algorithm first generates a random number $a_k$ from a uniform distribution of [0, 1]. $a_k$ is used to decide whether to use exploration or exploitation. If $a_k > \varepsilon$, then the exploitation strategy is adopted and the node with the highest action value in $S$ is selected as $Nod_k$. Otherwise, a node in $S$ is randomly selected as $Nod_k$ through the exploration strategy. What's more, if $a_k > 0.5$, the connection type is concat: $Type_k = concatenation$. Otherwise, the connection type is residual: $Type_k = residual$.

**State Update**

$$S = S \cup Nod_k, \quad if\ k \leq t \quad (3)$$

The $S$ is updated based on the selected node $Nod_k$. If $Nod_k$ is not a predefined termination node, the new node $Nod_k$ is added to the $S$. If $Nod_k$ is a terminating node, the $S$ remains unchanged. $t$ is the index of the termination node.

$$U = U \cup Nod_k \quad (4)$$

$$\Upsilon = \Upsilon \cup Type_k \quad (5)$$

The $U$ is updated based on the selected action type $Type_k$ and action $Nod_k$. The actions of each step (i.e., connection type and target node) are added to $U$ and $\Upsilon$ respectively to record the actions throughout the sampling process. Υ and $U$ together form the connected samples. Specifically, $S$ contains the starting node of the connection, while $U$ contains the target node and Υ contains the connection type. For each starting node and its corresponding target node, a unidirectional connection is established.

**1.3) Generative network:** In order to analyze the performance of new connections, we need to put them to original network model for evaluation. The details are as follows:

**Graph Expansion:** Firstly, integrate the new connections represented by $S$, $U$, and Υ into the original graph $G$ to form the $G''$.

**Model Construction:** Secondly, transform the graph $G''$ into a trainable network model. Iterates through all nodes in the computational graph $G''$, we construct layers in a deep learning framework (e.g., TensorFlow or PyTorch) according to each node type and attributes. Additionally, activation functions, batch normalization, and dropout layers can be added based on the node attributes. The constructed layers are then assembled into a model $M'$, ensuring that the graph is topologically sorted to respect layer dependencies.

Importantly, during the above model construction process, it is essential to consider the upstream and downstream relationships of the newly added connections (represented by $S$, $U$, and Υ) in $G''$. This is because the target layer of new connections may already have other existing connection inputs. Specifically, we analyze the source layer and target layer of each edge in $G''$. If necessary, adjustments are made by adding layers such as $1 \times 1$ convolutional layers, fully connected layers, or reshape layers to match the requirements.

**1.4) Evaluation network:** To evaluate the performances of neural network model $M'$, we utilized a quantitative assessment method. Model $M'$ was trained for a restricted cycle, typically around ten epochs, to facilitate an early performance review while conserving computational effort. This is essential in estimating the model's prospective impact in practical settings. Employing this technique allows for a prompt and efficient appraisal of the capabilities of model $M'$, serving as a foundation for further improvements.

**1.5) Update Q values:** The update of the Q-table follows the Bellman equation, which is applicable to the sequence states $S$ and actions $U$. The update process is as follows:

**The update of the Q values:** For the last state-action pair in $S$ and $U$ directly depends on the accuracy of the currently evaluated model.

$$Q(S_{-1}, U_{-1}) = (1 - \alpha) \cdot Q(S_{-1}, U_{-1}) + \alpha \cdot accuracy \tag{6}$$

This step integrates the latest accuracy directly into the Q table, reflecting the impact of recent actions on model performance.

**Application of Bellman's equation:** For state-action pairs $(S_i, U_i)$ in the sequence, the $Q$ value is updated to take into account the immediate reward r and the maximum expected return under the next state $S_{i+1}$. The iterative update formula is modeled as follows:

$$Q(S_i, U_i) = (1-\alpha) \cdot Q(S_i, U_i) + \alpha \cdot (r + \gamma \cdot max(Q(S_{i+1}),\cdot)), i \in \{1,2,\ldots,M\} \quad (7)$$

Where $r$ is the calculated immediate reward and $max(Q(S_{i+1}))$ is the maximum action value for the next state. This update is based on Bellman's equation and reflects the long-term benefits of the current action.

In summary, as shown in **Fig. 3** (In the manuscript), the steps of connection samples to update Q values form a loop body, which is used to optimize the Q table and identify optimal new connections.

**2) Multi-scale dispersion entropy**

Brain signals exhibit complexity across multiple temporal scales, marked by long-range correlations and nonlinear behaviors. The application of MsDE [22] is critical for extracting these features, offering a measure of complexity at various scales and highlighting underlying patterns within neural activities. Inspired by MsRNN [3]. The calculation of MsDE involves: 1) Coarse-graining the original signal into different temporal scales. 2) Computing the dispersion entropy in the coarse-grained signal. Coarse-graining is performed by binning the original signal $Z = \{z_1, z_2, \ldots, z_t\}$ and calculating the mean value inside the bins of length $'s'$ to generate a new series $Y = \{y_1, y_2, \ldots, y_{\frac{\Phi}{\sigma}}\}$:

$$y_j = \frac{1}{\sigma}\sum_{k=(j-1)\sigma+1}^{j\sigma} z_k, 1 \leq j \leq \frac{\Phi}{\sigma} \quad (8)$$

The length $\sigma = 2$ of the bin, which is called the scale factor. $\Phi$ is the number of time points of the original signal $Z$. Next, to calculate the dispersion entropy of each time series of $Y$, we use the normal cumulative distribution function (NCDF) to map each element of the series into one of $c$ classes (a label $e_i$ whose value ranges from 1 to $c$). This mapping is pivotal for quantifying the complexity of the time series. The specific formula is as follows:

$$\theta_i = \frac{\sqrt{2\pi}}{D} \int_{-\infty}^{z_i} \exp\left(-(z_i - avg)/2D^2\right)dx \quad (9)$$

followed by,

$$e_i = \text{round}\,(c^*\theta_i + 0.5) \quad (10)$$

By fixing the mean $avg$ and standard deviation $D$ of the signal $Y$, the NCDF method utilizes a sliding window of length $\beta$ and step size $\tau$ to count the frequency of various $c^\beta$ dispersion patterns within the signal, and calculates the Shannon entropy of these frequencies to quantify the complexity of the signal. This approach reveals that the complexity of the signal reaches its maximum when the signal is completely random, due to the equal probability occurrence of all possible dispersion patterns; conversely, when the signal exhibits periodicity, showing only a single dispersion pattern, its complexity is minimized [31, 32]. For the parameters setting, we adopted the values of $c = 6, \tau = 1, \beta = 2$ as recommended in [31, 32]. The multi-scale dispersion entropy is $E = \{e_1, e_2, \ldots, e_{\frac{\Phi}{\sigma}}\}$.

3) **Multi-scale dilated convolution**

Considering insights from [10] that highlight the partially ordered nature of brain network transitions, it becomes pertinent to enhance the existing model by incorporating these dynamics. However, traditional multiscale convolution captures multiscale information by deploying convolution kernels of different sizes within the same layer [3] or by stacking different levels of convolution operations to improve the robustness of the model to scale variations. Such an approach fails to capture the complex, non-linear interactions between non-adjacent brain regions effectively. Therefore, we proposed a multi-scale dilated convolution (MsDC), which achieves feature extraction at different scales by adjusting the dilation rate, aiming to further enhance the feature capture capability of the model, especially when dealing with tasks with wide scale variations. A multi-scale dilated convolution is adopted to significantly enhance the model's ability to capture multi-scale spatial information by combining convolutional layers with different dilation rates. Each dilated convolution layer corresponding to a dilation rate independently expands the receptive field to integrate features from different spatial scales. This multi-scale dilation rate structure enables the model to efficiently handle input features at different scales, enhancing the ability to understand. Incorporating MsDC would be beneficial as it allows for capturing relationships between non-adjacent brain regions and reflects partial order dynamics of the brain. This approach has the potential to deepen our understanding of schizophrenia-related changes in brain activity and improve accuracy in classifying schizophrenia.

In our model, the MsDC is strategically applied within two crucial components: the multi-scale GRU block and the Residual Dilation block, as depicted in **Fig. 2 (In the manuscript)**. Within the residual dilation block, MsDC is utilized with different dilation rates to expand the receptive field across various spatial scales. This setup aids in analyzing brain interactions more comprehensively, enhancing the model's sensitivity to subtle variations in brain network dynamics. Within the multi-scale GRU block, MsDC pre-processes the input data, using varying dilation rates to enhance spatial feature extraction. This allows the GRU layer to handle richer temporal dynamics, improving the detection of intricate brain activity patterns related to schizophrenia. By incorporating MsDC into these blocks, our model better captures complex, non-linear interactions between non-adjacent brain regions, crucial for understanding and classifying schizophrenia.

4) **Transformer fusion module**

The utilization of the transformer module for feature fusion is driven by its ability to learn relationships among multi-feature vector effectively. By leveraging a transformer architecture, multi-feature extracted by the encoder can be comprehensively understood and enhance the classification performance. As illustrated in **Fig. 1** (In the manuscript), the transformer-based fusion module mainly consists of multi-feature vectors concatenation, a self-attention mechanism, a multi-head attention mechanism (MHSA) and feed-forward neural networks (FNN) [27].

**4.1) Multi-Feature Vectors Concatenation.** The feature data of each type is processed by its corresponding feature extractor to generate a feature vector, and these vectors are then concatenated. For instance, if the concatenation involves $n$ feature vectors, it is considered as a sequence consisting of $n$ tokens. Specifically, we input FNC $x \in R^{B \times B}$, TC $z \in R^{\Phi \times B}$, dFNC $p \in R^{(\frac{\Phi-W}{S}+1) \times \frac{B(B-1)}{2}}$, and Multi scale dispersion entropy $e \in R^{\frac{\Phi}{\sigma} \times B}$ to specific-model to obtain feature vector $\{\omega^1, \omega^2, \ldots, \omega^i\}$ for each sample respectively, where $v^i \in R^{1 \times \varphi}$ represents the different 'token' for the transformer, $i = 4$. $\varphi$ is the dimension of features. This design enables the Transformer model to effectively discern the interconnections among different features, thereby significantly improving the accuracy of classification tasks.

**4.2) Self-Attention Mechanism.** Each token can interact with any other tokens [33], allowing it to learn the dependencies among multiple features. Set $g = \{\omega^1, \omega^2, \ldots, \omega^i\}$, we map the input $g$ to new space through 3 different linear layers, resulting in 3 matrices, namely query, key, and value as follows:

$$Q = gW^Q, K = gW^K, V = gW^V \qquad (11)$$

where $W^Q \in R^{B \times d^q}, W^K \in R^{B \times d^k}, W^V \in R^{B \times d^v}$ are learnable linear matrices and $Q \in R^{B \times d^q}$, $K \in R^{B \times d^k}$, $V \in R^{B \times d^v}$ denote query, key and value respectively. Note that each row of query $Q^i$, key $K^i$ and value $V^i$ represent the deep features of $i^{th}$ token. In this study, we set $d^q = d^k = d^v = 4 \times \varphi$. To measure the similarity of the $i^{th}$ feature query $Q^i \in R^{1 \times d^q}$ to all other features on the key $K$, we computed the dot-product between them: $S_{ij} = Q^i K^{j^T}$. Then we scaled $S_{ij}$ by $\sqrt{d}$ and applied a softmax function to normalize it [27, 34], resulting in $\Gamma_{ij} = softmax(S_{ij}/\sqrt{d})$, which is the attention weights and indicates dependency between the $i^{th}$ feature and the $j^{th}$ feature. Each feature can interact with other features through the transformer encoder, thus capturing the global correspondence of the whole brain. Finally, the corresponding representation for $i^{th}$ feature $Q^i$ was obtained by assigning the attention weights with value $V$: $O^i = \sum_j \Gamma_{ij} V^j$ [34]. In summary, the output of self-attention mechanism can be formulated as follows:

$$O = \text{Softmax}\left(\frac{QK^T}{\sqrt{d}}\right) V \qquad (12)$$

**4.3) Multi-Head Attention Mechanism (MHSA).** To enrich the diversity of feature subspace, MHSA was proposed to project the input $g$ into different feature subspaces [35]. Each subspace was fed into a SAM module in parallel. The outputs of the self-attention mechanism modules were concatenated and project into a single feature with the same dimension as the input $g$. In summary, the process of MHSA is defined as follows:

$$MHSA(LN(x)) = \text{Concat}(O_1, O_2, \cdots, O_h)W^O \qquad (13)$$

where $h$ is the number of independent self-attention mechanism, $LN$ denotes the layer normalization operation, Concat is the concatenation operation, $W^O$ the parameter of output projection and $O_i$ is the output of $i^{th}$ SAM.

**4.4) Feed-Forward Neural Networks (FNN).** FNN consists of two linear layers with a GeLU activation, where 512 and 256 hidden nodes are used for each layer respectively. The model is trained using cross entropy losses.

# Results

This section consists of four experimental modules: 1) Multi site pooling classification. 2) Leave-one-site-out Classification. 3) Most Discriminative Independent Components and FNC. 4) Ablation Experiments.

**A. Implementation Details**

In our study, we assessed the model performance using ten-fold cross-validation. We randomly divided subjects into ten groups, using one as the test set and the rest for training in each iteration. We calculated the mean and standard deviation of various performance metrics, including accuracy (ACC), specificity (SPE), sensitivity (SEN), F1-score (F1), and AUC, to ensure the stability and accuracy of the evaluation. Two-sample t-test was employed to compare the performance of various algorithms.

The encoders of the proposed BRIEF were optimized using the Adam optimizer, with a learning rate set to 0.001 and dropout set to 0.5, which was further adjusted with a decay rate of 0.01, all implemented within the Keras framework. A batch size of 64 was consistently used throughout the training process. To manage weight sparsity in the multi-scale GRU block, we incorporated Dropout at a rate of 0.5 and applied L1 and L2-norm regularization with parameters L1 = 0.0001 and L2 = 0.0001 to the GRU layers. For the transformer-based fusion module, the learning rate is set to be 0.005 with a decay rate of 0.001. For the brain-inspired NCS: 1) $\varepsilon$-greedy Strategy. The agent is trained using Q-learning with experience replay [36] and $\varepsilon$-greedy strategy [37]. With epsilon-greedy strategy, the random action is taken with probability $\varepsilon$ and the greedy action is chosen with probability $1-\varepsilon$. We decrease epsilon from 1.0 to 0.1 following the epsilon schedule, where the number of iterations varies as follows: for ε = 1.0, the algorithm converges in 100 iterations; for ε = 0.9, 0.8, and 0.7, it takes 7 iterations; for ε = 0.6, 0.5, 0.4, 0.3, and 0.2, it stabilizes at 10 iterations; and for ε = 0.1, it increases to 12 iterations. This allows the agent to transition smoothly from exploration to exploitation, indicating potential thresholds in the algorithm's convergence behavior. The result improves with longer exploration, as it expands the search scope and allows the agent to discover more connections during the random exploration period. 2) Experience Replay. Following [38], we employ a replay memory to store the validation accuracy and candidate connections after each iteration. 4) network generation. In the Q-learning update process, the learning rate $\alpha$ is set to 0.01 and the discount factor $\gamma$ is 1. The initial learning rate is set to 0.001 and is reduced with a factor of 0.2 every 5 epochs.

**B. Multi-site pooling classification**

To demonstrate the advantages of BRIEF, we conducted comprehensive multi-site pooling experiments with ten-fold cross-validation, utilizing fMRI data aggregated across all sites. These experiments were iterated ten times to ascertain mean values and standard deviations for the evaluation metrics. BRIEF's performance was benchmarked against 21 approaches on both the In-House and ABIDE datasets. The quantitative outcomes of classification tasks are detailed in Table 2 and 3, and represented in **Fig.** 4. We compare BRIEF with a large family of baselines, including 21 models.

**FNC-based Methods.** Given the important applications of traditional machine learning in brain disease diagnosis, we compare BRIEF with the following machine learning methods: 1) Random Forest, 2) AdaBoost, and 3) SVM. We also compare with several popular deep learning models: 1) BrainNetCNN [39], which leverages the topological locality of brain networks through edge-to-edge, edge-to-node, and node-to-graph convolutional filters for age prediction; 2) Siamese GCN [40], which takes into consideration the graph structure for the evaluation of similarity between a pair of graphs; 3) Triplet GCN [41], which extracts multi-scale graph representations of brain FC networks; 4) Population-GCN [42], which involves representing populations as a sparse graph, and phenotypic information is integrated as edge weights based on GCNs; 5) A-GCL[6], which employs graph contrastive learning with edge-dropped graphs from trainable Bernoulli masks to extract discriminative features.

**TC-based Methods.** 1) GRU, which consists of two stacked GRU layers and multi-layer perception; 2) C-MLP, which consists of three convolutional layers and multi-layer perception; 3) SC-RNN, which consists of single convolutional layer, two stacked GRU layers and multi-layer perception; 4) Ms-RNN [5], which takes advantage of the high-level spatiotemporal information through multi-scale recurrent architectures; 5) ACIFBN [9], which addresses limitations in traditional functional brain network construction by modeling asynchronous functional interactions for improved diagnosis; 6) iTransformer [7], which inverts the roles of attention mechanisms and feed-forward networks to better capture multivariate correlations in time series data; 7) PatchTST [8], which segments time series into subseries-level patches as transformer input tokens, significantly improving long-term forecasting accuracy.

**DFNC-based Methods.** 1) LSTM, which consists of two stacked LSTM layers and multi-layer perception; 2) Full-BiLSTM [43], which learns the periodic brain status changes using both past and future information and then fuse them to form the final output; 3) GRU, which consists of two stacked GRU layers and multi-layer perception; 4) C-LSTM, which consists of three Conv1D layers, two LSTM layers, and multi-layer perception for classification.

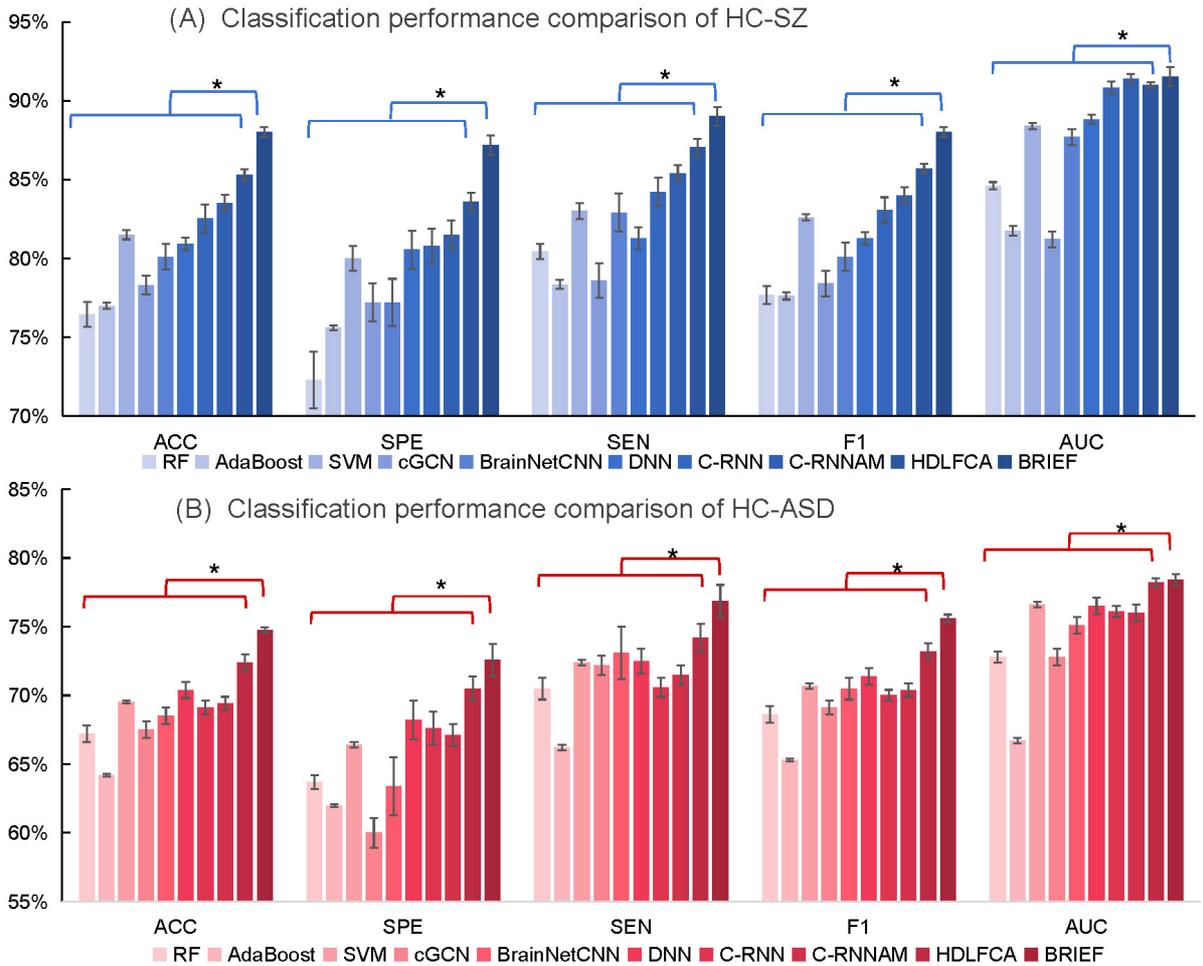

**Figure. 4** Comparative experimental performance. (A) Classification Performance Comparison on the SZ In-House dataset. (B) Classification Performance Comparison on the ABIDE dataset. The methods marked in green, blue and yellow color represent the FNC-based methods, the TC-based methods and the methods using both TC and FNC. * Denotes that the proposed BRIEF method achieves significantly better performance than the baselines with P value=0.05/0.01. The results showed that the proposed BRIEF outperformed all baselines using either static FNC or TCs or both.

As shown in **Fig. 4**, in our multi-site pooling classification experiment, we designed experiments encompassing four distinct features: TCs, FNC, dFNC, and MsDE. We employed both advanced machine learning and deep learning methods for comparative analysis. On the in-house dataset, our proposed method achieved an average accuracy (ACC) of 88.0% ± 0.3%, specificity (SPE) of 87.1% ± 0.6%, sensitivity (SEN) of 89.0% ± 0.6%, F1 score of 88.0% ± 0.3%, and area under the curve (AUC) of 91.5% ± 0.6%. In comparison to the SOTA results, these metrics exhibit improvements of approximately 2.8%, 3.6%, 2.0%, 2.4%, and 0.5%, respectively. The results on In-House dataset demonstrated remarkable superiority of our proposed method. Similarly, on the AIBDE dataset, our method excelled with an ACC of 74.7% ± 0.2%, SPE of 72.5% ± 1.2%, SEN of 76.8% ± 1.2%, F1 score of 75.5% ± 0.3%, and AUC of 78.4% ± 0.5%. Our method recorded enhancements of approximately 2.5%, 2.1%, 3.2%, 2.4%, and 0.2% in ACC, SPE, SEN, F1 score, and AUC, respectively, compared to the previous SOTA performance on the AIBDE dataset. This consistent uplift across multiple evaluation metrics solidifies the establishment of a new classification standard.

**Table 2** Performance comparison in multi-site pooling classification on In-House schizophrenia datasets

| Methods | Feature | ACC | SPE | SEN | F1 | AUC |
|---|---|---|---|---|---|---|
| RF | FNC | 75.9±1.2 | 72.6±2.2 | 80.4±0.5 | 77.6±0.5 | 84.2±0.4 |
| AdaBoost | FNC | 76.5±0.4 | 74.8±0.5 | 78.3±0.6 | 77.6±0.4 | 81.8±0.4 |
| RFE_SVM | FNC | 79.0±0.6 | 79.3±1.0 | 80.5±1.3 | 81.6±0.6 | 87.5±0.5 |
| BrainNetCNN[39] | FNC | 78.6±1.2 | 79.2±1.8 | 78.0±2.2 | 78.8±1.3 | 86.3±1.2 |
| Siamese GCN[40] | FNC | 77.8±1.1 | 78.8±1.8 | 76.5±1.7 | 78.2±1.1 | 86.0±1.0 |
| Triplet GCN[41] | FNC | 77.4±1.0 | 78.4±2.2 | 75.9±1.8 | 78.0±0.8 | 85.7±0.9 |
| Population-GCN[42] | FNC | 80.0±0.8 | 78.2±1.5 | 81.9±2.1 | 80.7±0.7 | 88.2±0.6 |
| A-GCL[6] | FNC | 77.6±1.3 | 78.2±2.1 | 74.1±1.8 | 77.3±0.9 | 84.8±0.7 |
| LSTM | dFNC | 80.5±0.5 | 81.5±1.2 | 79.6±1.0 | 80.6±0.5 | 88.8±0.3 |
| Full-BiLSTM | dFNC | 81.2±0.8 | 81.4±2.2 | 79.8±1.9 | 80.9±1.2 | 89.5±0.9 |
| GRU | dFNC | 80.6±0.9 | 80.5±1.1 | 81.1±2.3 | 81.1±1.2 | 88.7±0.6 |
| C-LSTM | dFNC | 79.6±0.7 | 80.2±2.0 | 78.9±1.2 | 79.7±0.6 | 88.0±0.4 |
| GRU | TCs | 76.9±0.5 | 74.4±1.0 | 79.3±0.7 | 77.8±0.5 | 84.3±0.3 |
| C-MLP | TCs | 77.1±0.4 | 75.7±0.8 | 78.4±0.7 | 77.7±0.4 | 86.7±0.3 |
| SC-RNN | TCs | 80.5±0.5 | 79.4±1.0 | 81.4±0.9 | 80.9±0.5 | 88.5±0.4 |
| ACIFBN[9] | TCs | 75.2±0.8 | 73.7±1.6 | 76.1±1.2 | 74.3±0.9 | 80.8±0.7 |
| iTransformer[7] | TCs | 80.3±0.7 | 79.8±1.4 | 82.1±1.1 | 81.3±0.8 | 87.1±0.7 |
| PatchTST[8] | TCs | 81.7±0.6 | 80.1±1.4 | 83.1±0.9 | 82.0±0.8 | 89.3±0.6 |
| MsRNN [5] | TCs | 82.2±1.2 | 80.2±1.5 | 83.8±0.8 | 82.5±0.7 | 90.1±0.6 |
| cGCN | FNC+TCs | 78.3±0.6 | 77.2±1.2 | 78.6±1.1 | 78.4±0.8 | 81.2±0.6 |
| HDLFCA[2] | FNC+TCs | 85.2±0.4 | 83.5±0.6 | 87.0±0.5 | 85.6±0.3 | 91.0±0.3 |
| **BRIEF** | **FNC+TCs+ dFNC+MsDE** | **88.0±0.3** | **87.1±0.6** | **89.0±0.6** | **88.0±0.3** | **91.5±0.6** |

**Table 3** Performance comparison in multi-site pooling classification on ABIDE datasets

| Methods | Feature | ACC | SPE | SEN | F1 | AUC |
|---|---|---|---|---|---|---|
| RF | FNC | 66.5±0.8 | 63.1±0.9 | 70.5±1.1 | 68.2±1.2 | 72.2±0.8 |
| AdaBoost | FNC | 64.0±0.5 | 62.2±0.6 | 65.8±0.6 | 65.1±0.5 | 66.3±0.6 |
| RFE_SVM | FNC | 67.9±0.5 | 66.5±0.6 | 71.9±0.8 | 70.2±0.7 | 75.8±0.7 |
| BrainNetCNN[39] | FNC | 67.5±0.6 | 63.4±2.1 | 73.1±1.9 | 70.5±0.8 | 75.1±0.6 |
| Siamese GCN[40] | FNC | 67.2±0.8 | 63.0±2.2 | 72.8±1.5 | 70.1±0.9 | 75.0±0.8 |
| Triplet GCN[41] | FNC | 67.3±0.8 | 62.9±1.8 | 72.5±1.8 | 70.2±0.7 | 75.2±0.7 |
| Population-GCN[42] | FNC | 69.2±0.6 | 66.2±1.5 | 72.1±1.2 | 70.0±0.9 | 76.3±0.6 |
| A-GCL[6] | FNC | 66.0±0.9 | 65.4±1.7 | 72.1±1.3 | 69.7±0.9 | 76.1±0.8 |
| LSTM | dFNC | 67.5±0.9 | 65.8±1.3 | 70.8±1.6 | 69.1±1.5 | 74.8±0.6 |
| Full-BiLSTM | dFNC | 68.0±1.0 | 67.2±2.1 | 70.5±1.8 | 69.9±1.8 | 75.3±1.0 |
| GRU | dFNC | 67.5±0.8 | 62.7±0.8 | 72.5±1.8 | 70.7±0.7 | 75.7±0.7 |
| C-LSTM | dFNC | 66.0±0.8 | 63.1±1.5 | 70.5±1.3 | 68.9±1.0 | 73.5±0.8 |
| GRU | TCs | 66.3±0.8 | 63.8±1.5 | 70.1±1.3 | 68.1±1.0 | 73.2±0.8 |
| C-MLP | TCs | 67.5±0.9 | 65.8±1.3 | 70.8±1.6 | 69.1±1.5 | 74.8±0.6 |
| SC-RNN | TCs | 68.2±1.0 | 66.9±2.1 | 70.1±1.8 | 69.6±1.8 | 75.3±1.0 |
| ACIFBN[9] | TCs | 65.9±1.1 | 63.7±1.9 | 69.8±1.7 | 68.1±1.6 | 73.1±0.9 |
| iTransformer[7] | TCs | 68.3±0.9 | 67.4±1.9 | 70.4±1.6 | 69.5±1.5 | 75.7±0.8 |
| PatchTST[8] | TCs | 68.7±0.8 | 67.5±1.8 | 70.5±1.5 | 69.9±1.6 | 76.1±0.8 |
| MsRNN [5] | TCs | 69.3±0.9 | 67.8±1.6 | 70.8±1.2 | 70.1±1.3 | 76.2±0.8 |
| cGCN | FNC+TCs | 67.0±1.2 | 60.0±1.3 | 72.4±1.5 | 69.5±0.9 | 72.5±0.9 |
| HDLFCA[2] | FNC+TCs | 72.2±0.6 | 70.4±0.9 | 73.6±1.0 | 72.9±0.6 | 78.2±0.3 |
| **BRIEF** | **FNC+TCs+ dFNC+MsDE** | **74.7±0.2** | **72.5±1.2** | **76.8±1.2** | **75.5±0.3** | **78.4±0.4** |

Furthermore, the combined evidence from Table 2, Table 3 and Fig. 4 bolsters the effectiveness of our BRIEF framework in exploiting multi-feature information. While Fig. 4 showcases the improved classification capability of our method when differentiating healthy controls (HC) from individuals with schizophrenia and autism spectrum disorder (ASD) using various feature combinations. In Table 3 and Table 4, through ablation studies, provides compelling evidence of the incremental contribution of each individual feature to the overall classification performance. This collective analysis underscores the presence of complementary information within brain temporal patterns, functional dynamics, dynamic functional dependencies, and multi-scale entropy. By judiciously integrating these diverse features, our model achieves a substantial enhancement in classification efficacy, thus validating the synergistic potential of these distinct neuroimaging features. In summary, our BRIEF approach, integrating a multi-scale network with extensive temporal features, consistently outperforms existing SOTA models in classifying brain disorders. This synergy harnesses complementary brain dynamics, as evidenced by superior performance across multiple metrics and datasets, demonstrating the value of jointly leveraging multi-scale architectures and multi-scale neuroimaging data for enhanced diagnostic precision in neurological conditions.

### C. Leave-one-site-out Classification

In the context of cross-site validation, we employed a rigorous leave-one-site-out strategy, where each site was iteratively withheld as a test set while the remaining sites were utilized for model training. To ensure model generalization and prevent overfitting, 10% of the subjects within the training portion were randomly selected for internal validation within the BRIEF framework. The performance of BRIEF was rigorously assessed on the In-House dataset, with the corresponding quantitative results detailed in **Table 4.**

Table 4 Performance comparison in leave-one-site-out classification between HC and SZ

| Methods | Feature | ACC | SPE | SEN | F1 | AUC |
| --- | --- | --- | --- | --- | --- | --- |
| RF | FNC | 71.4±4.4** | 65.1±1.2** | 79.1±8.1 | 73.5±4.0* | 82.1±3.5** |
| AdaBoost | FNC | 74.8±2.5** | 74.1±6.7** | 75.7±4.7 | 75.1±3.3* | 82.1±2.6** |
| SVM | FNC | 77.2±3.6* | 76.6±9.7** | 78.5±6.5 | 77.6±4.0 | 85.5±4.4** |
| BrainNetCNN | FNC | 75.8±3.8* | 77.5±9.5** | 75.8±6.3 | 76.5±3.2 | 85.1±4.2** |
| DNN | FNC | 76.8±3.1* | 76.2±9.0** | 77.8±5.7 | 77.8±3.7 | 85.0±4.0** |
| C-RNN | TCs | 77.6±1.9* | 77.9±8.1** | 77.1±9.3 | 77.3±4.2 | 86.5±2.4** |
| C-RNN | TCs | 78.9±2.1 | 80.0±6.5* | 77.9±7.8 | 77.8±3.0 | 87.2±2.1* |
| cGCN | FNC+TCs | 75.1±3.2* | 76.5±9.0* | 74.7±5.6 | 75.5±3.2* | 83.1±4.1** |
| HDLFCA | FNC+TCs | 81.5±2.2 | 87.5±6.0 | 75.1±5.8 | 80.3±1.7 | **90.2±2.4** |
| **BRIEF** | **FNC+TCs+dFNC+MsDE** | **85.8±0.9** | **88.4±1.4** | **83.3±2.6** | **85.4±1.2** | 89.8±1.8 |

The BRIEF exhibited a pronounced increase in diagnostic precision compared to a variety of established machine learning and deep learning algorithms, including BrainNetCNN, Deep Neural Network (DNN), Convolutional Recurrent Neural Network (C-RNN), Attention-augmented C-RNN (C-RNNAM), cGCN and HDLFCA. This enhanced performance substantiated the value of the BRIEF in automatically extracting and exploiting high-dimensional fMRI features for schizophrenia (SZ) classification. The BRIEF achieved higher accuracy than these competing methods, with improvements ranging from 14.4% to 4.3%, thereby demonstrating its superior ability to discern SZ-related patterns within fMRI data.

A careful examination of **Table 2, 3** revealed that the BRIEF consistently outperformed cGCN and HDLFCA, further underscoring its exceptional ability to extract and exploit the intricate patterns of functional connectivity and dynamic neural activity embedded within fMRI data. This superior performance demonstrates the BRIEF's capacity to effectively discern and capitalize on disease-relevant information across multiple scales, leading to enhanced classification efficacy in the context of schizophrenia (SZ). In summary, the BRIEF emerged as a robust and efficacious instrument for deciphering the intricacies of SZ-related brain dysfunction, contributing significant advancements to the domain of neuroimaging-based diagnostics through its innovative and comprehensive approach.

## D. Most Discriminative Independent Components and FNC

In this section, we employed attention maps to analyze the discriminative power of independent components (ICs) and their functional network connectivities (FNCs). For schizophrenia (SZ [44-46]), the attention analysis identified six key brain regions ranked by their discriminative importance: striatum, precentral gyrus, cerebellum, inferior frontal gyrus, cerebellum, and superior frontal gyrus [30-32]. The connectivity analysis revealed that the most discriminative functional connections occurred between subcortical networks (SC) and visual networks (VIS), sensorimotor networks (SMN) and visual networks (VIS), frontal networks (FN) and attention networks (ATN), as well as connections within attention networks (**Fig. 5A, B**).

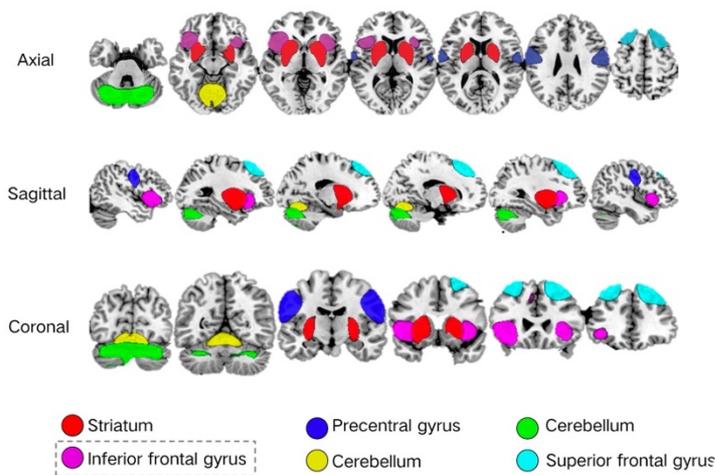
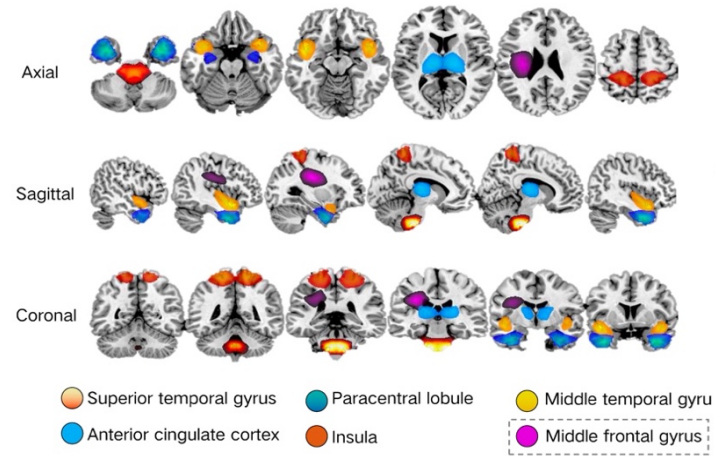
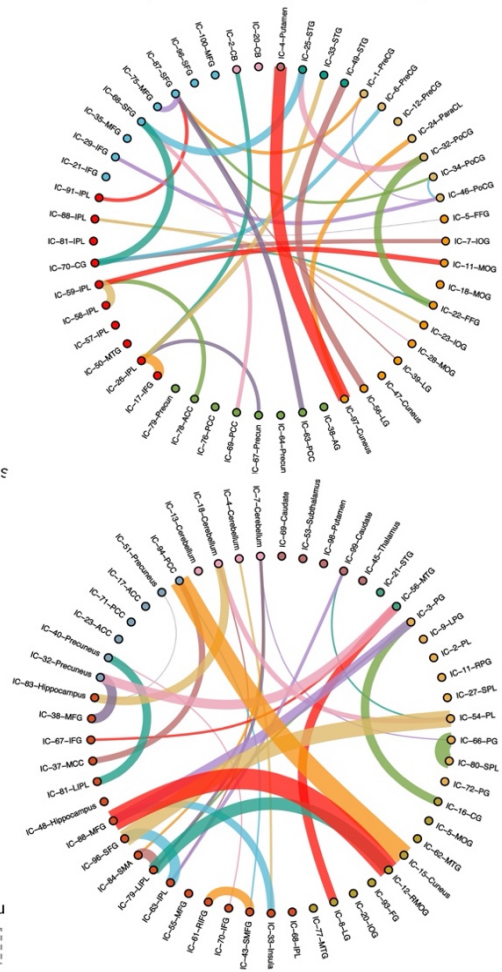

**Figure. 5** Top contributing brain networks and functional connectivities identified by attention method in schizophrenia (SZ) and autism spectrum disorder (ASD). (A) The 6 most important ICs in SZ, including striatum, precentral gyrus, and cerebellar regions. (B) Top 30 functional network connections in SZ, highlighting interactions between subcortical, visual, sensorimotor, and attention networks. (C) The 6 most important ICs in ASD, including temporal, cingulate, and frontal regions. (D) Top 30 functional network connections in ASD, showing key interactions between visual, cognitive-control, default-mode, and sensorimotor domains.

For autism spectrum disorder (ASD), a different set of brain regions emerged through the attention analysis, with the following importance ranking: superior temporal gyrus, paracentral lobule, middle temporal gyrus, anterior cingulate cortex, insula, and middle frontal gyrus. The most discriminative functional connections for ASD were identified between visual (VI) and cognitive-control (CC) domains, visual and default-mode (DM) domains, sensorimotor (SM) and cognitive-control domains, and within the sensorimotor domain (**Fig. 5C, D**).

Taken together, these visualizations underscore the distinctive connectivity patterns that differentiate SZ and ASD. SZ appears to involve significant disruptions in subcortical and attention-related connectivity, while ASD is marked by unique alterations in sensory integration, cognitive control, and default mode connectivity. These distinct patterns of regional involvement and network connectivity highlight the different neural characteristics associated with SZ and ASD.

**E. Ablation Experiments**

For feature component ablation studies as shown in **Table 5**: 1) FNC + TCs: This base configuration integrates Functional Connectivity (FNC) with Time Series (TCs), showing robust performance across all evaluation metrics. 2) FNC + TCs + dFNC: With the addition of dynamic functional connectivity, the model demonstrated slight improvements in ACC, SPE, SEN, and F1, despite a small drop in AUC. 3) FNC + TCs + dFNC + MsDE: The inclusion of multi-scale Dispersion Entropy (MsDE) to the existing configuration significantly enhanced all performance indicators, strongly affirming the importance of extensive temporal features in the classification of schizophrenia.

Table 5 Feature ablation studies on in-house schizophrenia datasets

| Feature | ACC | SPE | SEN | F1 | AUC |
|---|---|---|---|---|---|
| **FNC+TCs** | 87.6±0.4 | 86.6±0.4 | 88.1±0.3 | 87.2±0.3 | 91.3±0.4 |
| **FNC+TCs+dFNC** | 87.9±0.5 | 86.8±0.5 | 88.4±0.3 | 87.8±0.3 | 91.0±0.4 |
| **FNC+TCs+dFNC+MsDE** | **88.0±0.3** | **87.1±0.6** | **89.0±0.6** | **88.0±0.3** | **91.5±0.6** |

For network component ablation studies as shown in **Table 6**: 1) w/o NCS: Removing the NCS led to a decline in performance metrics, demonstrating its importance in enhancing the model's classification capabilities. 2) w/o Multi-dilated Convolution: Omitting multi-dilated convolutions showed a slight performance drop, underscoring the effectiveness of multi-scale dilated convolution in extracting diverse scale features. 3) LR Fusion, w/o Transformer Fusion: A decrease in performance was observed when the model did not include a Transformer-based feature vector fusion module, further validating the significance of this advanced feature vector fusion module based on Transformer architecture. 4) Proposed BRIEF: By comparison, our complete BRIEF framework, incorporating all features and network components, achieved the best performance, validating the synergistic effect of combining extensive temporal features and advanced network architecture components.

Table 6 Method ablation studies on in-house schizophrenia datasets

| Methods | ACC | SPE | SEN | F1 | AUC |
| --- | --- | --- | --- | --- | --- |
| w/o NCS | 86.1±0.3 | 86.3±0.5 | 87.1±0.4 | 86.0±0.3 | 90.5±0.3 |
| w/o MsDC | 88.0±0.4 | 86.7±0.6 | 88.7±0.5 | 87.8±0.4 | 91.1±0.2 |
| w/o Transformer | 86.5±0.4 | 86.0±0.8 | 86.8±0.7 | 86.1±0.3 | 90.2±0.3 |
| **Ours** | **88.0±0.3** | **87.1±0.6** | **89.0±0.6** | **88.0±0.3** | **91.5±0.6** |

Note: */** denote that the proposed BRIEFmethod achieves significantly better performance than the baselines with P value=0.05/0.01.

In summary, the ablation study distinctly highlighted the crucial roles of extensive temporal features and network components in enhancing the classification performance on a schizophrenia dataset. Our experiments demonstrated that each component—whether it be the integration of Functional Connectivity and Time Series, the addition of dynamic functional connectivity, or multi-scale Dispersion Entropy—contributes significantly to the model's effectiveness. Moreover, the network architecture, especially the inclusion of brain-inspired network connection search, multi-dilated convolutions, and Transformer-based feature vector fusion, proved essential for optimizing classification capabilities. Overall, these findings validate the efficacy of the BRIEF, offering valuable insights for future research and clear directions for ongoing model optimization.

# Discussion

We proposed a BRIEF framework that combined extensive temporal features and network to classify brain disorders, where multi scale network can distill the complementary and robust high-level representations of extensive temporal features (FNC/dFNC/TCs/MsDE). Specifically: 1) We have developed a universal connection optimization method called NCS to optimize state-of-the-art network flow connections. Additionally, we used multi-scale dilated convolutions to extend the multi-scale convolutional method [3]. These components together enhance the feature extraction capability of multi-scale networks. 2) By deeply analyzing brain functional connectivity and activity, this study successfully integrates extensive temporal features and networks, offering a new perspective for improving the classification of brain disorders. 3) The Transformer model further improved classification performance for ASD and schizophrenia by integrating multiple feature vectors extracted by multi-scale networks. After the above optimization, our proposed BRIEF achieved an accuracy of 88.0% and 74.7% in distinguishing healthy controls (HCs) from SZ and ASD across multiple sites, comprehensively surpassing the classification performance of the previous SOTA method.

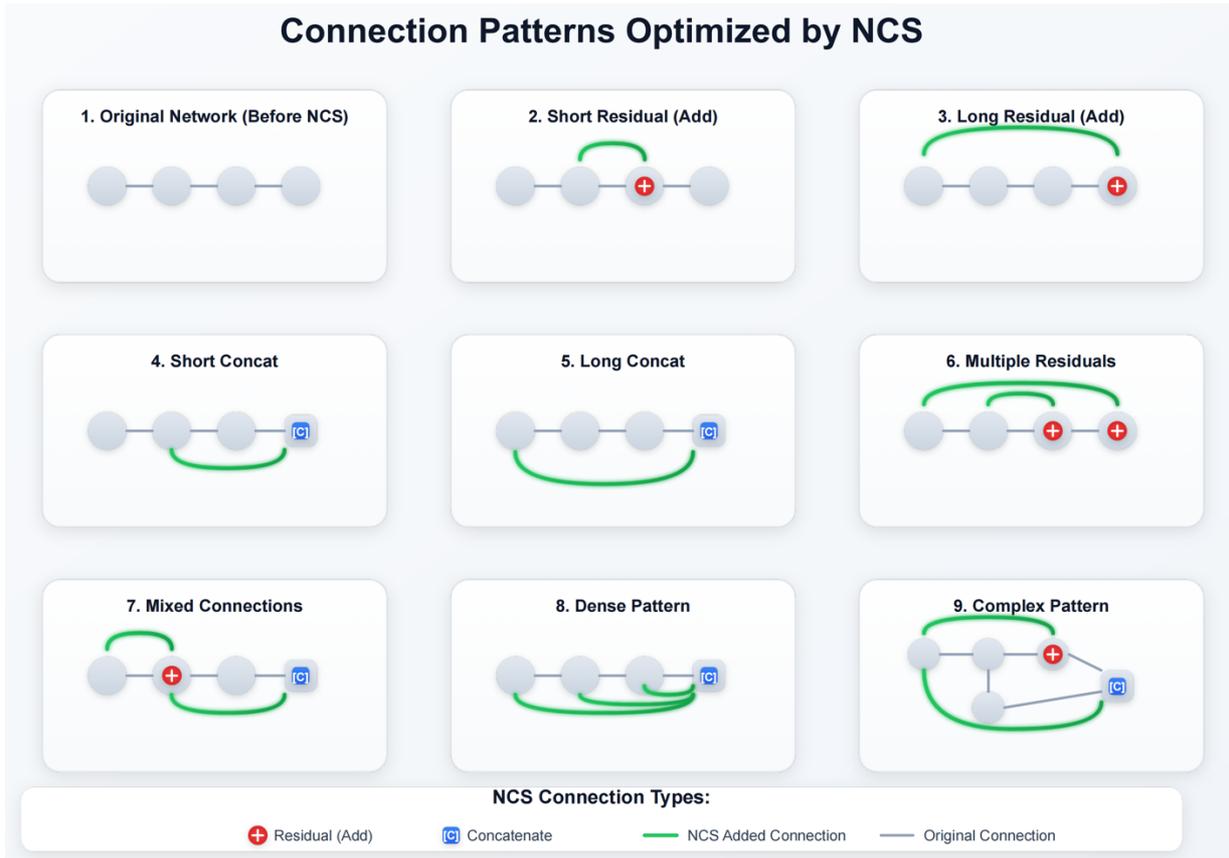

**Figure. 6** Diverse Connection Patterns Optimized by Neural Connection Search (NCS).

**NCS promotes network optimization** with the following insights: 1) We can optimize based on advanced networks rather than starting from scratch, as in NAS. NCS allows for faster convergence and better performance, leveraging the strengths of advanced models. By refining existing networks and incorporating neuroscientific insights, the NCS provides a promising avenue for enhancing the efficiency and accuracy of neural networks. This approach not only accelerates the discovery of new architectures but also addresses some of the computational and scalability challenges associated with traditional NAS methods. 2) The connections between neurons in the brain are far more complex than those between different neurons in neural networks (deep learning). Therefore, network optimization should place importance on the connections. Specifically, NCS uses utilization of both long and short-range residual connections and concatenation connections. These types of connections have been shown to improve the accuracy of deep learning models by facilitating more effective information flow through the network. 3) Inspiration can be drawn from brain science research. Although there may be significant differences between brain intelligence and modern neural networks, there are also many similarities. The collaborative development of brain science and artificial intelligence is beneficial to both fields.

As shown in **Fig. 6**, this figure presents mutiple neural architectures discovered by the Brain-Inspired Neural Connection Search (NCS), highlighting its ability to reproduce established designs (e.g., ResNet-style residuals in Panels 2~3, DenseNet-like dense concatenations in Panel 8) while generating novel biologically plausible topologies (Panels 6~7, 9). Aligned with Section 4.1, NCS refines advanced networks rather than building from scratch—green paths denote Q-learning-optimized connections that enhance feature flow through short/long-range residuals (red "+" in Panels 2-3) and multi-scale concatenations (blue "[C]" in Panels 4~5, 8). Hybrid patterns (e.g., Panel 7's residual-concat mix) emulate the brain's hierarchical connectivity (Section 4.2), such as corticostriatal pathways (long residuals in Panel 3) and default-mode interactions (dense concats in Panel 8). By bridging neurobiological principles with computational efficiency, NCS achieves superior classification (SZ: 91.5% AUC), validating its role in unifying connection search.

**NCS provides inspiration for Reasoning**. In current large language models (LLMs) learning strategy, using the Transformer and leveraging large computational power to exploit the capabilities of scaling laws [47] has endowed large models with strong reasoning abilities. However, since strict reasoning is not considered during model training, the application of reinforcement learning (RL) in models currently serves mainly as an alignment mechanism, which can degrade the performance of models when dealing with complex problems. Recent advances in reasoning architectures have demonstrated that alternative approaches to traditional CoT methods can achieve superior performance on complex reasoning tasks with significantly fewer parameters [48]. Therefore, we believe that the insight from proposed NCS is that RL can be utilized to optimize the reasoning process, particularly in long and short path search and decision-making mechanisms, elevating the model's reasoning capabilities to a higher level.

**Potential way to optimize network connections using NCS.** In fact, we are optimizing network connections based on a SOTA network and have set the final number of connections to be added. If you want to maximize the use of NCS to optimize network connections, one possible solution is to design a base network that contains as few connections as possible, and then hand over to NCS to optimize the remaining candidate connections, and set the target number of connections as high as possible, then NCS will play a greater role in automatically optimizing network connections. The efficiency of optimization will be much higher than that of manual design.

**Brain Components Sorting.** Furthermore, we integrated an attention mechanism into encoder to identify key brain networks and generate attention maps highlighting the most influential independent components (ICs). Analysis revealed that the striatum, precentral gyrus, inferior frontal gyrus, cerebellum, and superior frontal gyrus are critical discriminative regions. Notably, the striatum—which includes the putamen and caudate—is significantly highlighted in schizophrenia research as part of the basal ganglia circuitry, affecting motor control, decision-making, and stimulus-response learning [49]. Previous studies consistently report alterations in the structure and function of the striatum in schizophrenia, including dopaminergic and glutamatergic dysregulation and neuroinflammation [50]. Moreover, functional connectivity involving the cerebellum, known for reduced white matter integrity and blood flow in schizophrenia, is also significantly correlated with the disorder [44, 45]. These findings confirm that our attention-based model effectively captures essential aspects of schizophrenia-related brain dysfunction, providing a new perspective on understanding the neural mechanisms of this complex condition.

**ABIDE Performance Analysis:** As is well known, the ABIDE dataset is a public dataset. However, the preprocessing and data selection processes vary across different studies, resulting in a notable distribution of accuracy levels. To ensure a fair comparison of algorithm performance, we have uniformly applied the open-source code from several studies to our dataset for comparison. The results demonstrate that our proposed BRIEF achieves optimal performance. Similarly, we also applied these open-source algorithms to our private dataset, where our BRIEF likewise achieved the best performance.

**Limitations**. Although proposed BRIEF enhances the expressiveness of features, we also identified some limitations. Specifically, there is a separation between the feature fusion module and the different feature extraction networks, which does not leverage the benefits of mutual learning. This separation may limit the model's performance in handling complex data structures. Future work could explore how to effectively facilitate interaction between these components to improve the overall performance of the model.

In conclusion, we present a significant advancement in neuroimaging analysis through the development of the Brain-Inspired Feature Fusion (BRIEF) framework, which innovatively addresses the challenges of feature extraction and integration in fMRI data processing. Our framework makes three key methodological contributions: First, we introduce a brain-inspired Network Connection Search (NCS) strategy that revolutionizes neural architecture optimization by leveraging reinforcement learning mechanisms, offering an efficient alternative to traditional Neural Architecture Search approaches. Second, we expand the temporal feature representation beyond conventional metrics to encompass time courses, static and dynamic functional connectivity, and multi-scale entropy, capturing richer neural dynamics. Third, we implement a transformer-based fusion mechanism that effectively integrates these diverse features for enhanced diagnostic accuracy. Comprehensive evaluation using large-scale, multi-site datasets of schizophrenia (n～1100) and autism spectrum disorder (n～1522) demonstrated the framework's exceptional performance, achieving accuracy improvements of 2.2-12.1% over 21 state-of-the-art models. Notably, our attention mechanism successfully identified disorder-specific brain regions that align with established neurobiological findings, enhancing the framework's clinical interpretability. These results not only validate the effectiveness of our approach but also demonstrate its potential as a reliable tool for objective psychiatric diagnosis. Looking forward, the BRIEF framework, powered by its novel NCS optimization strategy, opens new avenues for both methodological advancement in deep learning and clinical applications in psychiatry. The successful integration of brain-inspired computational principles with advanced machine learning techniques suggests promising directions for developing more sophisticated and interpretable diagnostic tools across various neurological and psychiatric conditions.

## Conclusion

In this study, we proposed BRIEF, a novel framework that combined extensive temporal fMRI features fusion with brain-inspired neural network search strategy. To the best of our knowledge, this is the first attempt to incorporate brain-inspired, reinforcement learning strategy to optimize fMRI-based mental disorder classification. We utilized four temporal features, including TC/sFNC/dFNC and MsDE, collectively reflecting time-varying fluctuations and long-range functional dependencies. To effectively extract high-level vectors from each of the four features encodes, we designed a smart network connection search (NCS) strategy inspired by the brain decision-making mechanisms, in which NCS is formulated as a Markov Decision Process where actions are potential modifications to network connections, and rewards are based on performance improvements. The extracted feature vectors from each encoder were further input into a Transformer-based fusion module. BRIEF was validated on multiple datasets, including schizophrenia (SZ, n~1100) and autism (ABIDE, n~1522), demonstrating superior performance with accuracy improvements of 2.2-12.1% over 21 latest alternative models, proving its effectiveness on enhancing the accuracy of psychiatric diagnoses by leveraging multi-scale time-varying functional information, suggesting huge promise for identifying potential neuroimaging biomarkers.

## CRediT Author Statement

Xiangxiang Cui: Conceptualization, Methodology, Validation, Writing – original draft. Min Zhao: Conceptualization, Methodology, Writing – original draft. Dongmei Zhi: Data preprocessing, Writing – original draft. Shile Qi: Visualization, Writing – review & editing. Vince D Calhoun: Writing – review & editing. Jing Sui: Supervision, Writing – review & editing.

## Declaration of Competing Interest

The authors declare that they have no known competing financial interests or personal relationships that could have appeared to influence the work reported in this paper.

## Data Availability

ABIDE data publicly available at https://github.com/preprocessed-connectomes-project/abide.

In-house SZ data available upon reasonable request.

## Code Availability

All the code will be available at https://github.com/AbnerAI/BRIEF after the paper is accepted.

## ACKNOWLEDGEMENTS

This work was supported by the Scientific and Technological Innovation 2030 - The Major Project of the Brain Science and Brain-Inspired Intelligence Technology (2021ZD0200500), the National Natural Science Foundation of China (62373062), and the Startup Funds for Talents at Beijing Normal University.